# The structure of Bayes nets for vision recognition

*John Mark Agosta*
*Robotics Laboratory*
*Stanford University*
*Stanford, CA 94305*
johnmark@coyote.stanford.edu

## I. The problem

This paper[1] is part of a study whose goal is to show the efficiency of a Bayes net to carry out model based vision calculations. [Binford et al. 1987] The network model is drawn up from the object's geometric and functional description to predict the appearance of an object. Then this net can be used to find the object within a photographic image of a scene. Many existing and proposed vision recognition techniques resemble the uncertainty calculations of a Bayes net. In contrast, though, they lack a derivation from first principles, and tend to rely on arbitrary parameters that we hope to avoid by our network model.

     The connectedness of the net depends on what independence considerations can be identified in the vision problem. Greater independence leads to easier calculations, at the expense of the net's expressiveness. Once this tradeoff is made and the structure of the net is determined, it should be possible to tailor a solution technique for it.

This paper explores the use of a net with multiply connected paths, drawing on both techniques of belief networks [Pearl 86] and influence diagrams. We then demonstrate how one formulation of a multiply connected net can be solved.

---

[1] This paper grew out of extensive discussions with Tom Binford, Dave Chelberg, Tod Levitt and Wallace Mann. I owe a special debt of gratitude to Tom for introducing me to both the problem and a productive way to approach it. As usual all errors are the sole responsibility of the author.

## II. Nature of the vision problem

The objects within a visual image offer a rich variety of evidence. The image reveals objects by their surface edges, textures, color, reflectance, and shadows. Save for extreme cases, only a part of this evidence is necessary to recognize an object.

The vision problem is no simpler because of the surfeit of evidence, since the use of evidence is problematic. The researcher may approach this preponderance of clues by concentrating only on a limited variety, such as that which is easiest to calculate, or leads to the most efficient algorithm. Bayes methods encourage the use of a wider variety of evidence since they traditionally have been developed to integrate diverse and subtle sources of evidence.

The problem of image recognition has close kin. By extension to the variety of evidence, an object could be identified by kinds not available in a visual image, such as its tactile feel or its motion when it is disturbed. Likewise there are vision problems for which recognition is not necessary, such as visual obstacle avoidance. Recognition is one aspect of understanding the scene; the scene may be analyzed to infer the location of the viewer relative to the object and to make other functional statements. Conceivably the process of recognition might proceed to a higher level recognition of some situation or "Gestalt." These concerns are outside the scope of this paper.

Model based vision consists of two activities; first, modelling the objects to be looked for – the predictive phase, then identifying objects by analyzing a raster image – the inferential phase. Vision models are built from a top down decomposition of an object's geometry, into geometric primitives that further decompose into primitive observable features. The object is decomposed into sub-assemblies that are in turn decomposed to obtain relations among volume filling primitives, in our case "generalized cylinders" [Binford 1971] and their intervening joints. Volumes have surfaces that generate shapes. The shapes are projected onto the image visual plane as regions defined by line segments and junctions. The lowest level features are the edges and patches they define.

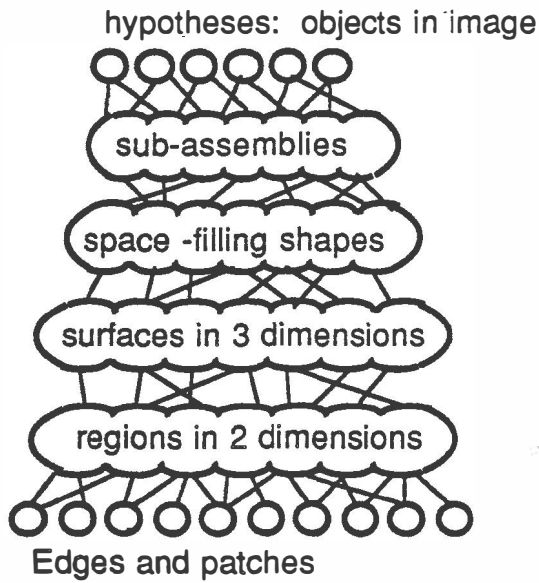

Vision recognition based on such a model proceeds by grouping image features at lower levels to identify features at the next higher level. The bottom-up grouping process is driven by a decomposition model built for each object expected to be within the image.

### III. Issues in formulation

Grouping is a primary process of recognition, and occurs at each level of the decomposition hierarchy. A predictive arc from a feature to a lower level feature implies both the appearance of a lower level feature and its location relative to the feature. If a predictive arc did not entail some position information, then the net could not perform grouping of features. Consider the contrary case: Complete but non-localized evidence of the number and kind of low level features that compose an object, such as pairs of lines intersecting at certain angles. This is still only weak evidence for the appearance of the object. Unfortunately grouping adds dependencies that complicate the net structure. Here I discuss the formulation of these dependencies.

*A. Tree structured hierarchies.*

The formulation of a decomposition model implies a probability net where top level constructs predict the appearance of lower level ones. When stages can be decomposed into independently observable parts, the net becomes tree structured. The tree is rooted (at the "top"; an unfortunately confusing use of terms) in a hypothesis about the appearance of an object. It grows "down" to leaf nodes that represent image primitives. The process of recognition begins when evidence from an image instantiates primitives. Then by inference from lower level nodes to higher levels, the calculation results in the probability of top level hypotheses – that these objects are in the view.

For example, a generalized cylinder consists of a face, an axis and a sweeping function for the face cross section along the path of the axis. The face of the generalized cylinder appears independently of the axis. The lower level observables of both face and axis remain independent. In contrast, the sweeping rule and the limbs that it predicts depend on both the face and axis.

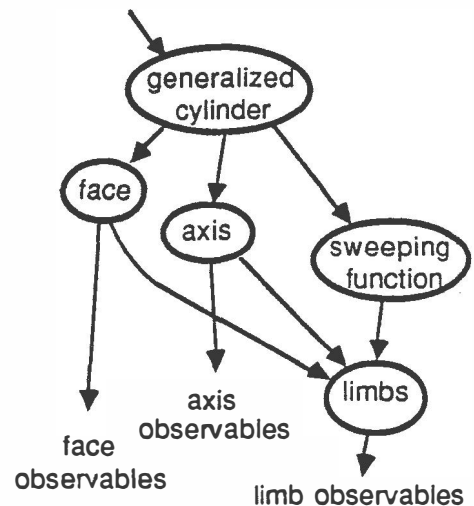

## B. Ambiguity

Optical illusions – images that appear to be more than one object– are an example of the ambiguious functioning of the human vision system. The novelty of such images indicates their rarity. This is evidence that in the eye, the process of bottom up grouping rarely resorts to backtracking. Ambiguity can be formulated as the existence of a higher level construct, which once determined, clarifies an ambiguity more than one level below. It is expressed by influences that skip over levels. They express kinds of arguments that stand in contrast to the grouping-at-each-level kind of reasoning. Hopefully, the intermediate constructs in the hierarchy are rich enough so that they are not necessary.

## C. Exclusion

Nodes with multiple parents in the net may be used to express exclusion. In this way evidence for one object implies the other does not exist.

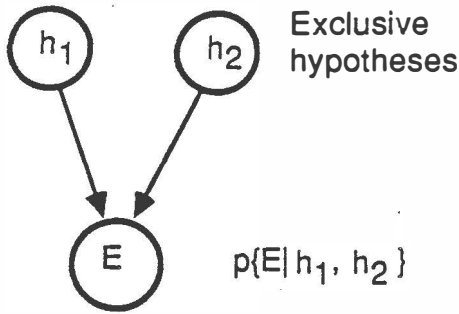

Exclusive hypotheses

$p\{E|h_1, h_2\}$

The evidentiary node for their exclusion is a sibling of both objects. Here instantiating E offers evidence for one hypothesis but not both. The distribution function of the evidence given the object hypotheses, $p\{E | h_1, h_2\}$, resembles an exclusive-or function. We can demonstrate the strong dependence between object nodes that this evidence generates by flipping the direction of the arc between a hypothesis and the evidence. By Bayes rule this generates an arc between hypotheses. Thus the truth of one hypothesis upon observing the evidence depends strongly on the truth of the other.

This characteristic multiple parent structure allows formulation by Bayes nets of conflicting hypotheses sets as presented by [Levitt 1985].

More useful for the purposes of formulation is an evidentiary node that excludes hypotheses for two objects at the same location. For this, the structure of the net is the same, but the meaning of the node differs.

## D. Co-incidence

Just as multiple parents can express exclusion, they can be used to infer the appearance of two models at the same location. Enforcing co-incident locations for diverse models may be a useful modeling tool. Imagine an object that could be posed as two separate models depending on the level of detail. For instance the "Michelin Man" could be modeled as both a human figure and as a stack of tires. As evidence for him, the perceptor would expect to find both a man and a stack of tires in the same location.

More common objects may also be composed as a set of co-incident models. For example a prismatic solid may be interpreted differently as generalized cylinders, depending on the choice of major axis. We may infer more than one of these generalized cylinders occupying the same location from which we infer the one object that predicted the set.

Composition of the same object as several co-incident models is not to be confused with the decomposition hierarchy. Decomposition is essentially a conditional independence argument, that the features into which an object is decomposed can be recognized separately, given the object appears in the view. Composition as several co-incident models, like exclusion formulations, depends on one lower level feature being predicted by multiple higher level features. Such multiple parent structures describe relations among their ancestors. In coincident models, furthermore, the multiple higher level features are resolved into the one ancestor that admits of co-incident interpretations. This generates a net with multiple paths.

*E. Global location and the use of proximity information.*

In general, of evidence for exclusion and co-incidence is evidence of the kind that predicts the proximity of two higher level constructs. For instance, when an observable feature in the image is an artifact of the joint between two features, then the proximity of the features may be inferred.

Further, all parts of the hierarchy are influenced by object position, orientation and articulation. All variables are a function of these global variables. They form a set of variables with universal influences, as shown here:

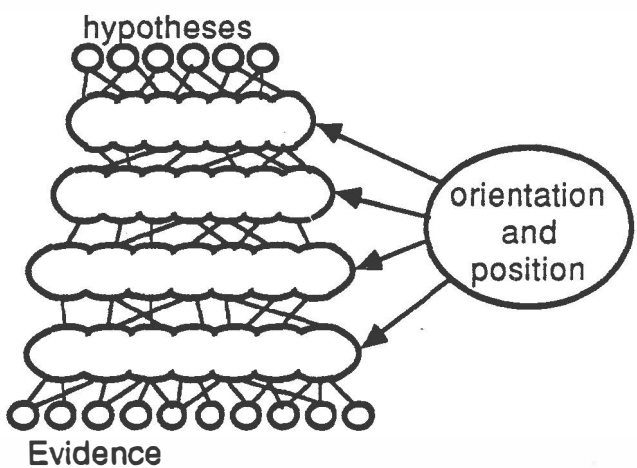

This generates a set of arcs that violate tree type hierarchies. As inference proceeds, orientation and position parameters form an intermediate hypothesis upon which top level hypotheses depend. These dependencies allow information about feature orientation or position to be made available to superior nodes, once an inference about them can be made from a lower level.

Again a question arises about the tradeoff between making this an explicit influence, or eliminating such global nodes and encoding "softer" proximity information within the net structure. Since the geometric modeling and the inference calculations that complete orientation information require are demanding, it is worth considering whether less specific proximity information could be used efficiently as a substitute. Arguably, a person can recognize an image with distorted orientations among components, much as the subject of a cubist painting with its orientation distortions can be recognized.

Thus, if as I have argued, the net cannot be completely formulated as a tree, it is likely that local relationships – exclusion, co-incidence and proximity can be exploited that do not require "global" influences. These local relationships resemble the traditional methods of vision recognition by a process of the repeated grouping of features.

**IV. Single verses multiply connected nets**

When more than one higher level feature jointly predicts one lower level observable within a recognition tree, it creates multiple connections. Fortunately suggested methods for vision model network formulation create networks with influences only between adjacent levels. This section presents a solution to a simple example that suggests a technique to apply to this restricted class of multiply connected nets.

*A. Concept of a solution*

Either influence diagram or belief net techniques result in the same solution to a directed acyclic Bayes net. Solving the net by influence diagram techniques derived from Bayes rule transforms the net so that a subset of the nodes, the set of hypotheses, is conditioned by the rest. The evidentiary nodes are tranformed so that they are not conditioned by other nodes. The initial distribution over the set of hypotheses, known as the prior, is contained in the net specification. If it was not explicit, e.g. the hypothesis nodes were not unconditioned originally, the prior of the hypothesis set can be found by taking expectation over all other nodes. Solving the net also imputes a distribution over the evidence, know as the pre-posteriors or marginals. The pre-posteriors - the distribution of evidence imputed by the priors before observations are made - has significance for the solution only as relates to further collection of information, etc. During solution, the hypothesis nodes' joint distribution changes as evidentiary nodes are instantiated. These distributions are replaced with degenerate distributions (e.g. observed values) and repeated application of "Jeffrey's rule" changes the hypothesis to distributions posterior to the evidence. These "posteriors" are the results of the solution, from which choices are made.

When solving by belief net operations, unconditioned distributions, or "beliefs" are maintained at all times for all nodes in the net. Initially, these represent the priors for the source nodes – the hypothesis – and the pre-posteriors for the evidentiary nodes – the leaves. As nodes are instantiated, message passing schemes update all nodes to their beliefs given the instantiations.

The subset of singly connected Bayes nets for which Pearl has derived a solution technique has the particular and useful property of being modular in space and proportional in time to the net diameter. There are several extensions for more complicated nets, by conditioning over a cut set [Pearl, 1985] and by star decomposition [Pearl 1986]. At the other extreme of complexity of directed acyclic probability graphs, [Shachter 1986] shows that there exists a solution to any influence diagram in a finite number of steps.

### B. Multiple hypothesis nets.

In the case when the various objects that appear in the field of view can be decomposed into a tree then these trees can be attached to a common set of leaf nodes. The result is no longer a tree, although it does not contain multiple directed paths. Is it possible that when the leaves are instantiated and are no longer probabilistic nodes, the trees, for purposes of solution, separate into a forest, and the calculation is equivalent to evaluating the trees sequentially? Then Pearl's algorithm could be applied directly to each tree. This conjecture is wrong. As Pearl recognizes, [Pearl 1985] nodes with multiple parents cannot be the separating nodes in a cutset. The parents of each leaf are not conditionally independent given the leaf node. This is apparent by application of Bayes rule through influence diagram transformations to condition the parent nodes upon the leaf node. Reversing a parent-to-leaf arc generates an arc between parents. This dependence is mediated in Pearl's algorithm by the message passing at the leaf nodes.

Another way to see the dependency is in terms of message passing. Incoming $\pi$s are reflected at leaves and affect the upward propagating $\lambda$s even when the belief at each leaf is certain, as shown by the propagation formula at an instantiated leaf node for $\lambda$s:

(1) $\lambda_i = \alpha \Sigma_j \pi_j p\{E_k | P_i P_j\}$

[Pearl 86, from equation 21] where $P_i$ is the parent receiving the lambda message, $P_j$ is the parent sending the pi message and k, an index over states of evidence, its value set by instantiation.

For separation it is necessary to assume that the P matrices at the leaf node are singular. A leaf with conditional distribution

(2) $p\{E | P_1, P_2\} = p\{E|P_1\}p\{E|P_2\}$

is equivalent to evaluating separate trees where the leaves contain the probabilities $p\{E|P_i\}$. Equivalently, this assumption makes the outgoing $\lambda$s in equation 1 independent of the $\pi$ value entering the leaf node.

In exchange for the complexity that common leaf nodes add to the computation, they provide the ability to express exclusion and co-incidence among hypotheses in which the trees are rooted. This is the significance of equation 1: The $p\{E | P_1, P_2\}$ matrices mediate the effect between parent nodes.

### C. A solution to a multiply connected net

It is a reasonable conjecture that some multiply connected nets lend themselves to time-efficient solution techniques. We offer one example where a multiply connected net can be solved by an extension of Pearl's technique. For this simple case consider a pair of common-leaved trees only one level deep, with two hypotheses and two evidentiary nodes, shown in the following diagram:

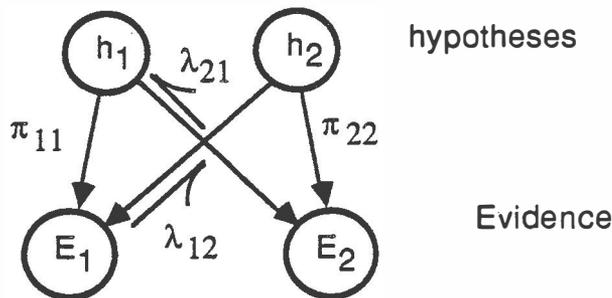

This net has a closed form solution in terms of Pearl's messages. The message sent by each node is a function of the message it receives. We derive these reflection equations from propagation rules at net boundaries. From the top-down propagation rule [Pearl 1986, p.260] for root nodes, the hypothesis sends down a $\pi_{HE}$ message vector that is the term by term product of its prior with the $\lambda$ message it receives:

(3) $\pi_{H\,E} = \alpha \pi_i \lambda_i$.

For each hypothesis node, write this as the product of a vector (all $\lambda$s and $\pi$s now represent vectors) and a diagonal matrix of the prior

(4) $\pi_{22} = \alpha \underline{h}_2 \lambda_{12}$
(5) $\pi_{11} = \alpha \underline{h}_1 \lambda_{21}$

where $\alpha$ normalizes $\pi$, and the underline meaning the matrix with the terms of $h$ on its diagonal.

From equation 1 reflection equations at the leaves are, in vector notation:

(6) $\lambda_{21} = \alpha P_2 \pi_{22}$
(7) $\lambda_{12} = \alpha P_1 \pi_{11}$

where, as before, the likelihood, $P_i = p(E_i | h_1, h_2)$, is indexed by which evidence, k, instantiates the node. So for each leaf node, the $\lambda$ propagating up is the matrix product of the likelihood with the $\pi$ it receives.

Substituting together (4)(5)(6) and (7) obtains

(8) $\pi_{11} = \alpha \underline{h}_1 P_2 \underline{h}_2 P_1 \pi_{11}$.

This expresses the cycle of messages in one direction. For the other cycle, containing $\pi_{21}$, the P matrix transposes appear in the corresponding leaf reflection equations.

Viewed as a recursion equation, if this has a solution, messages converge to steady values. Dividing through by the normalization constant puts this in the form of an equivalent eigenvector problem for the matrix $A = \underline{h}_1 P_2 \underline{h}_2 P_1$ with eigenvalue $1/\alpha$. This eigenvector is the solution to the recursion relation. It may be obtained directly, without recursion. Since all matrix elements are positive, it can be shown that the maximum eigenvalue and its eigenvector will be positive. From this eigenvector, the beliefs of all nodes can be calculated. For example, let the evidence at both leaves produce the same P matrices, and both $\underline{h}$ functions represent a uniform prior. Then

$P = \begin{bmatrix} .52 & .18 \\ .08 & .22 \end{bmatrix}$   $Bel(h_1) = \begin{bmatrix} 0.811 \\ 0.190 \end{bmatrix}$

$\alpha = 0.562$   $Bel(h_2) = \begin{bmatrix} 0.345 \\ 0.655 \end{bmatrix}$

$\pi = \begin{bmatrix} 0.811 \\ 0.190 \end{bmatrix}$

Further simplifications occur in the fortunate case when the net is a singly connected tree. Since the beliefs at intermediate nodes are not needed to solve for hypothesis posteriors, we need only propagate upward. Downward propagation is not necessary. More generally, in a net with multiple connections near the root, we can similarly ignore downward propagation in the singly connected extents of the tree, and propagate up in one pass to all nodes below multiple parent nodes. As before, determining the resulting beliefs (unconditioned probabilities) of evidentiary nodes is uninteresting as far as hypotheses are concerned. This simplification follows from the lambda updating formula [Pearl 86, p.260]. As this shows, $\lambda$s in a single parent node depend only on the $\lambda$s arriving from below, and not on the node's belief, nor on $\pi$s descending the tree. This is not symmetric with propagation in the other direction; the $\pi$s interact with the $\lambda$s on their way down. Thus, in a true tree, the hypothesis posterior can be updated solely by lambda propagation upward. This is equivalent to so-called "naive Bayes" updating schemes.

### V. Further directions.

Empirical tests will determine whether the directions described in this paper improve the ability of vision based modeling systems. There are also a host of formulation and solution questions to pursue.

What is the relation of the attachment graph – the graph of geometric relations – to the graph that predicts observable features (the "recognition net" that has been discussed here)? They are related, but not in a direct way. A further question this raises is the automatic generation of an recognition net from the objects' geometric description. As formulation questions are explored, can we avoid a fully connected recognition graph, approximate certain influences and still make efficient enough use of information for vision?

As we consider scenes with a wider variety of objects and the recognition engine gains flexibility by having more models from which to chose, the recognition net becomes bushier. A flexible recognition system may rely on control strategies for partial evaluation of the net.

Perhaps the value and decision structure that an influence diagram techniques add to the probability net evaluation suggest other solution techniques, in inference control, or hypothesis generation?